\documentclass[runningheads,a4paper]{llncs}

\usepackage{amssymb}
\usepackage{graphicx}

\usepackage{url}
\urldef{\mailsa}\path|{alexander.hagg, frederik.hegger, paul.ploeger}@h-brs.de|
\newcommand{\keywords}[1]{\par\addvspace\baselineskip
\noindent\keywordname\enspace\ignorespaces#1}

\begin{document}

\mainmatter  

\title{On Recognizing Transparent Objects in Domestic Environments Using Fusion of Multiple Sensor Modalities}

\titlerunning{Recognizing Transparent Objects in Domestic Environments}

%
%
\author{Alexander Hagg%
\and Frederik Hegger\and Paul G. Pl\"{o}ger}
\authorrunning{Recognizing Transparent Objects in Domestic Environments}

\institute{Bonn-Rhein-Sieg University of Applied Sciences, Department of Computer Science\\
Grantham-Allee 20, 53757 Sankt Augustin, Germany\\
\mailsa\\
\url{http://www.h-brs.de}}

\toctitle{Blabla}
\tocauthor{Alexander Hagg}
\maketitle

\begin{abstract}
Current object recognition methods fail on object sets that include both diffuse,  reflective and transparent materials, although they are very common in domestic scenarios.
We show that a combination of cues from multiple sensor modalities, including specular reflectance and unavailable depth information, allows us to capture a larger subset of household objects by extending a state of the art object recognition method. This leads to a significant increase in robustness of recognition over a larger set of commonly used objects.
\keywords{object recognition, transparency, fusion, modalities, domestic robotics, multimodal}
\end{abstract}

\section{Introduction}\label{sec:Introduction}

Object recognition is dominated by methods developed for color and depth cameras. Applications usually involve diffuse, textured materials. Environments in service robotics, as opposed to its industrial counterpart, tend to be much harder to control, increasing the need for high system robustness, whilst minimizing cost. The rise of low cost RGB-D sensors has begun to cut a path toward the latter goal but come with some old and new deficiencies. In this work we explore techniques to improve generalization and robustness with respect to realistic domestic environments and a wide set of objects, such as those that are transparent and cannot be described by the Lambertian reflectance model \cite{Lambert1760}. Such materials reflect large numbers of features from the environment and render common visual features unusable. Perception of transparent, reflective and refractive materials is one of the main problems that has not yet been solved in an affordable and generalizable way.

We combine a broader range of sensor modalities, which all have their benefits for certain material properties, similar to the categorization work by Marton et al. \cite{Marton2009b}, for visual object recognition. Modalities are defined as a transformation, $\mathcal{M}: (I \rightarrow V) \rightarrow f(V)$, with $I$ the image space, $V$ the sensor's value range, and $f(V)$ the transformation onto an interpretation of the sensor's values. The transformation represents a hypothesis of a certain physical property in the scene based on evidence by certain sensor values.

We extend a state of the art approach that allows for multimodal inputs and use the sensor's weaknesses to our advantage. The depth sensor returns \textit{nan} values for pixels for which no depth value is found, used to hypothesize a reflective or refractive material. We also provoke a saturated intensity response for specular materials by adding a light from a fixed position near the camera.

\begin{figure}[ht!]
\centering
	\includegraphics[width=60mm]{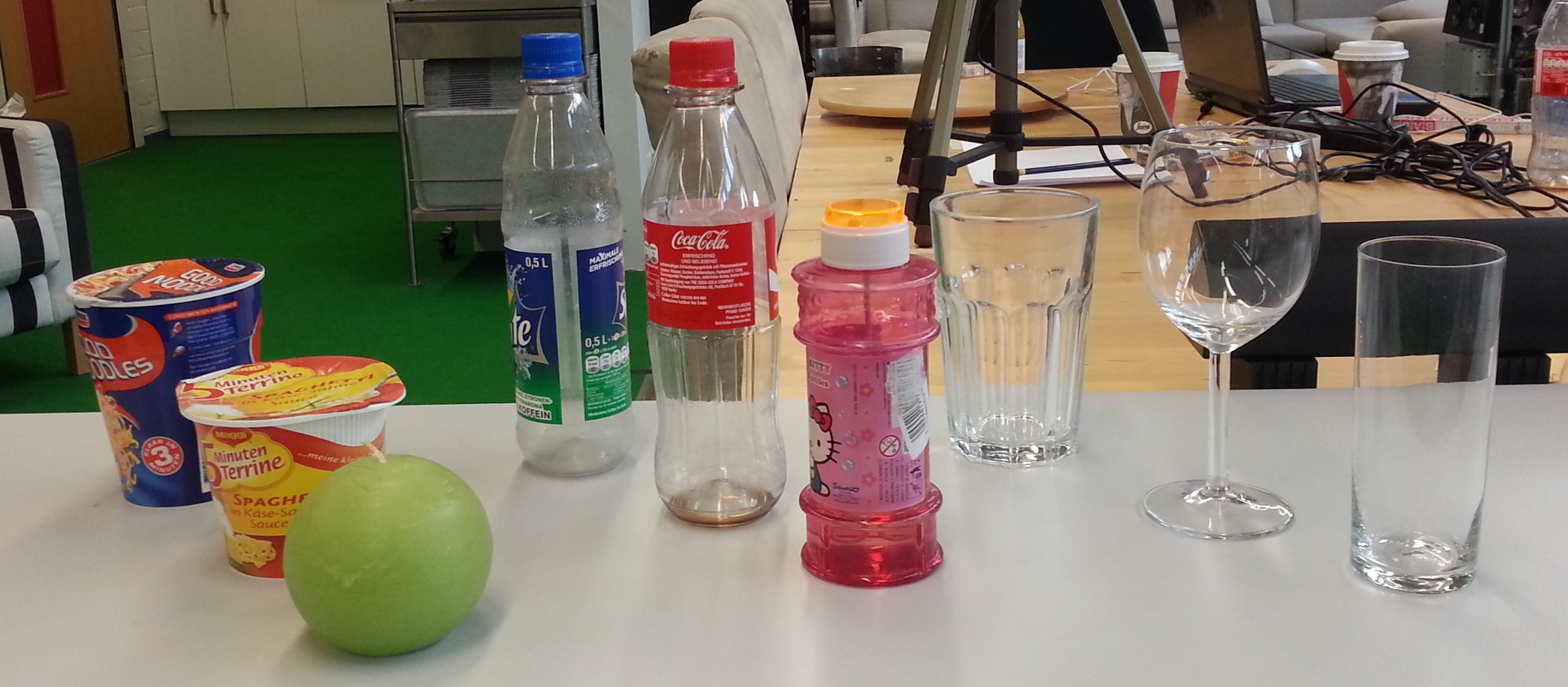}
	\caption[Evaluation set]{Object evaluation set with three categories of objects: \textit{diffuse}, \textit{composite} and \textit{transparent}.}
	\label{objects}
\end{figure}

We evaluate the proposed approach on three main object categories: \textit{diffuse textured}, \textit{semi-transparent} and \textit{composite}. The latter consists of a number of different materials from the first two categories. This set of materials covers most domestically used objects. Examples can be taken from Figure \ref{objects}. We compare our approach to the baseline system, LineMOD \cite{Hinterstoisser2012}, and analyze the used modality spaces and sensor characteristics. Experiments are run in a standard tabletop scenario, assuming that most household objects are found on horizontal surfaces such as tables and cupboards.

In the next section, we will discuss modalities that specialize on transparent materials. Section \ref{sec:RelatedWork} describes existing approaches and provide a more in depth insight into the approach we extended. In Section \ref{sec:Approach} we describe the modalities we used to increase the set of recognized objects, after which we evaluate our approach in the next section.

\section{Related Work}\label{sec:RelatedWork}

Most object recognition methods assume object sets whose visual response can be described by the semi-Lambertian reflectance model. However, reflective and transparent materials are described based on the specular reflectance model \cite{heath1981history}. Material-specific methods only serve a particular material or assume a controllable environment. Current approaches include measuring the polarization of light in highlights or using the refraction of a known background pattern to reconstruct a transparent surface. But these often require either full control of environmental lighting or a full model of the object and the environment behind it, which both do not apply to domestic environments.

A number of methods is based on the fact that specular reflection can cause light polarization changes. Koshikawa et al. \cite{Koshikawa1987} showed that these changes can be used to infer local surface normals. Saito et al. \cite{Saito1999} applied this technique on specular reflectance highlights. Others used similar approaches, also based on the near infrared (NIR) spectrum, but were mostly hampered by tight illumination constraints \cite{Zhang2012a,Mahendru2012}.
Fritz et al. \cite{Fritz2009} formulate a Latent Dirichlet Allocation \cite{Blei2003} in combination with SIFT \cite{Lowe2004}, which describes patch appearance based on the local edge energy distribution of refraction (caused by the underlying material). Maeno et al. \cite{Maeno2013} use a light field camera and model distortion by refraction. Both methods assume that the background has sufficient texture.
Albrecht et al. \cite{Albrecht} and Lysenkov et al. \cite{Lysenkov2013} use the observation that NIR structured light cameras are not able to produce depth data for transparent and most reflective materials, as the light is scattered away. Their approaches require a prior full 3D model.
Blake et al. \cite{Blake1991} describe and use a principle that is based on a priori knowledge of the position of a dominant light-source and inferring features from specular highlights caused by object materials that adhere to the specular reflection model. Object recognition methods from this approach always assume either prior model knowledge or fully controllable illumination. Klank et al. \cite{Klank2011} depend on large camera movements. Their method provides a slightly better than random guess on whether an object is transparent or not, not accounting for unexpected occlusion in one of the viewpoints. Albrecht et al. \cite{Albrecht} use unavailable depth data from an RGB-D sensor to reconstruct transparent objects. Wang et al. \cite{Wanga} improve transparent object segmentation using unavailable depth data as well. Alt et al. \cite{Alt2013}) use unavailable depth data to enhance object borders from multiple view points. Both specular highlights and unavailable depth data from an RGB-D camera fulfil the requirements within an affordable robotics context.

Chiu et al. \cite{Chiu2011} focus on improving perception of geometrical data by fusing multiple modalities and thus possibly allowing for the better acquisition of non-transparent data. The approach does not focus on adding information on transparent objects but instead incorporates \textit{missing information}, helping segmentation and localization of transparent objects.


Marton et al. \cite{Marton2009b} fuse sensor data from an RGB-D, a time of flight and a thermal sensor on a low level basis, although their categorization accuracy was only around 23\% for glasses. Their framework focusses on probabilistic categorization and is not extensible to individual recognition tasks that enable localization and grasping. Another low level multimodal object recognition systems, LineMOD, was introduced by Hinterstoisser et al. \cite{Hinterstoisser2012}. Their approach allows extension to other modalities and focusses on the recognition and localization task. We extend it and use it as a baseline system for evaluation. 

\section{Multimodal Approach}\label{sec:Approach}

LineMOD, defined by Hinterstoisser et al. \cite{Hinterstoisser2012}, is used because of its low level multimodal model. The authors define a novel low level abstract template representation for cues from any modality. Their approach is based on locally dominant gradient orientation for features, requiring that a feature is representable as such. As we will show in subsection \ref{subsec:Approach:Modalities}, the low level internal representation of LineMOD allows a wide range of modalities to be used. In common robotics scenarios, new objects are prone to appear often and household situations are subject to many user-introduced variations. By using quantization and spreading of bit-coded features, fast online learning whilst keeping generalization and robustness as high as possible makes the system fit for these scenarios. 

The authors use complementary modalities, compensating for each other's weaknesses. We too want to add modalities in a complementary, decoupled way to enhance the recognition system towards other reflectance models without degrading the performance of the original modality set. The authors of LineMOD use two modalities: \textit{maximum intensity gradients} to detect edges from desaturated RGB data and \textit{maximum normal vectors} from depth data to detect surfaces from diffuse objects. As we seek to describe features as patches of pixels showing certain local behaviours in various modality spaces, we are able to describe edges of such patches using the same dominant gradient approach as is used in LineMOD.

\subsection{Modalities}\label{subsec:Approach:Modalities}

By introducing a larger set of modalities, we provide a richer multimodal input that takes into account both semi-Lambertian and specular reflectance models. Table \ref{modalities_table} shows an overview of all used modalities. The modality $\mathcal{M}_1$ is based on \textit{maximum intensity gradients}. It is used to describe an object’s contour, as the original authors focus their recognition system on texture-less objects, where the foreground/background intensity difference is a good cue for the object's edge in the image. The gradients are calculated for each color channel to remove the influence of the object's and background's absolute color. This method is discriminant enough to describe texture as well, allowing the modality to describe both object edges as well as surface texture. The gradients are normalized in order to add robustness to contrast changes. The dot product between a template and the observed normalized vectors is used as a similarity measure.

\begin{table}[ht!]
\centering
	\caption[Modalities used in this approach]{Modalities used in this approach}
	\label{modalities_table}
		\begin{tabular}{|l|l|l|l|}
		\hline 
		modality	   &   physical property   &   $f(V)$	&	range\\
		\hline
		$\mathcal{M}_1$   &   2D shape   &   max. intensity gradients	&	$[0,\pi]$\\
		\hline 
		$\mathcal{M}_2$   &   3D geometry   &   max. normal vectors	&	$[0,\pi]^{2}$ \\
		\hline 
		$\mathcal{M}_3$   &   transparency   &   unavailable depth	&	$\{0,1\}$\\
		\hline 
		$\mathcal{M}_4$   &   specular reflection   &   max. intensity 	&	$\{0,1\}$\\
		\hline 
	\end{tabular}
\end{table}

\textit{Dominant normal vectors} from depth data, $\mathcal{M}_2$, based on NIR disparity images from the RGB-D camera, serve as a cue to the 3D shape of the visible surface. The features are defined as a least-square optimal gradient for a patch neighbourhood around the current pixel in the depth image. As a similarity measure, again the dot product between the template and perceived image serves as a similarity measure.

\begin{figure}[ht!]
\centering
	\includegraphics[width=50mm]{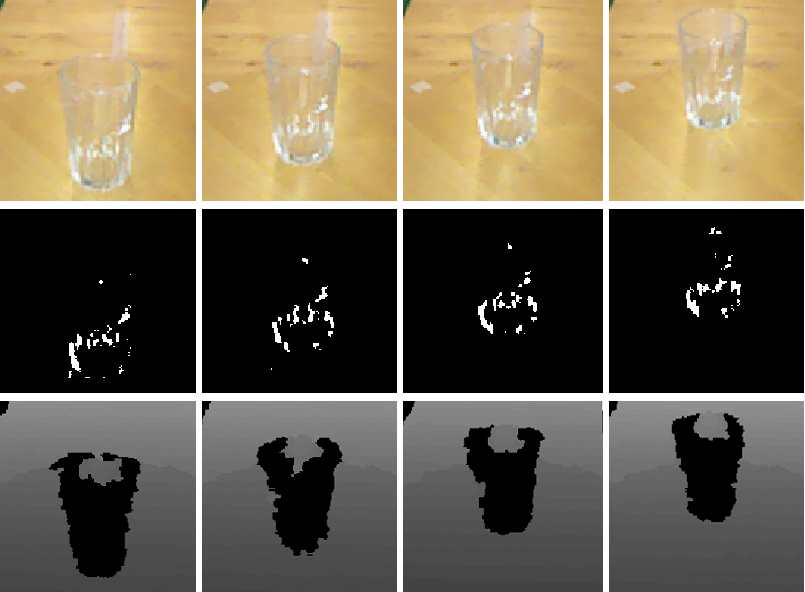}
	\caption[Changes in specularities in $y$ direction]{Water glass in various positions (+7cm. each) from front to back. \textbf{Top}: raw RGB input, \textbf{middle}: thresholded intensity images showing specularity candidates, \textbf{bottom}: visible shape from unavailable depth data from NIR camera (black).}
	\label{intensity-specularities-y}
\end{figure}

In addition to the modalities $\mathcal{M}_1$ and $\mathcal{M}_2$ we introduce two modalities $\mathcal{M}_3$ and $\mathcal{M}_4$. These modalities will allow the recognition of (semi-)transparent objects. As was already described by Albrecht et al. \cite{Albrecht}, the NIR pattern from the active RGB-D camera is reflected away from the camera or is irreversibly deformed by transparent objects. Large patches of unavailable depth data (observed as \textit{nan} values in the depth image) are observed for transparent and reflective objects, which is shown in the bottom of Figure \ref{intensity-specularities-y}. This can serve as a cue for the existence of a transparent material. It is not sufficient to accurately describe the material but does allow us to capture the 2D contour of the object, as can be seen in the lower part of Figure \ref{intensity-specularities-y}. We can use the same dominant gradient feature as was used for $\mathcal{M}_1$.

As many household objects tend to be generally symmetric around the vertical axis and this approach assumes a tabletop scenario, we can interpolate parts of the object’s 3D geometry by extruding the part of the contour where the object meets the table. We then use the interpolated 3D shape to find locally dominant normal orientations as was done for modality $\mathcal{M}_2$.

The second modality $\mathcal{M}_4$ we add to the pool is based on specular reflections, as in \cite{Zhang2012a}, which can be extracted by setting an absolute lower threshold to the intensity image. The specular reflections, caused by bundling of the environment light by the geometry of the object, were shown to be pose-invariant under the assumption of a dominant light by Netz et al. \cite{Netz2011}. Obviously, according to the laws of optical reflection, reflections of the environment by an object’s surface are not invariant. We therefore introduce an off-the-shelf 20 lux LED light at a fixed position near the sensors to add an invariant lighting factor into the scene.

\begin{figure}[ht!]
\centering
	\includegraphics[width=50mm]{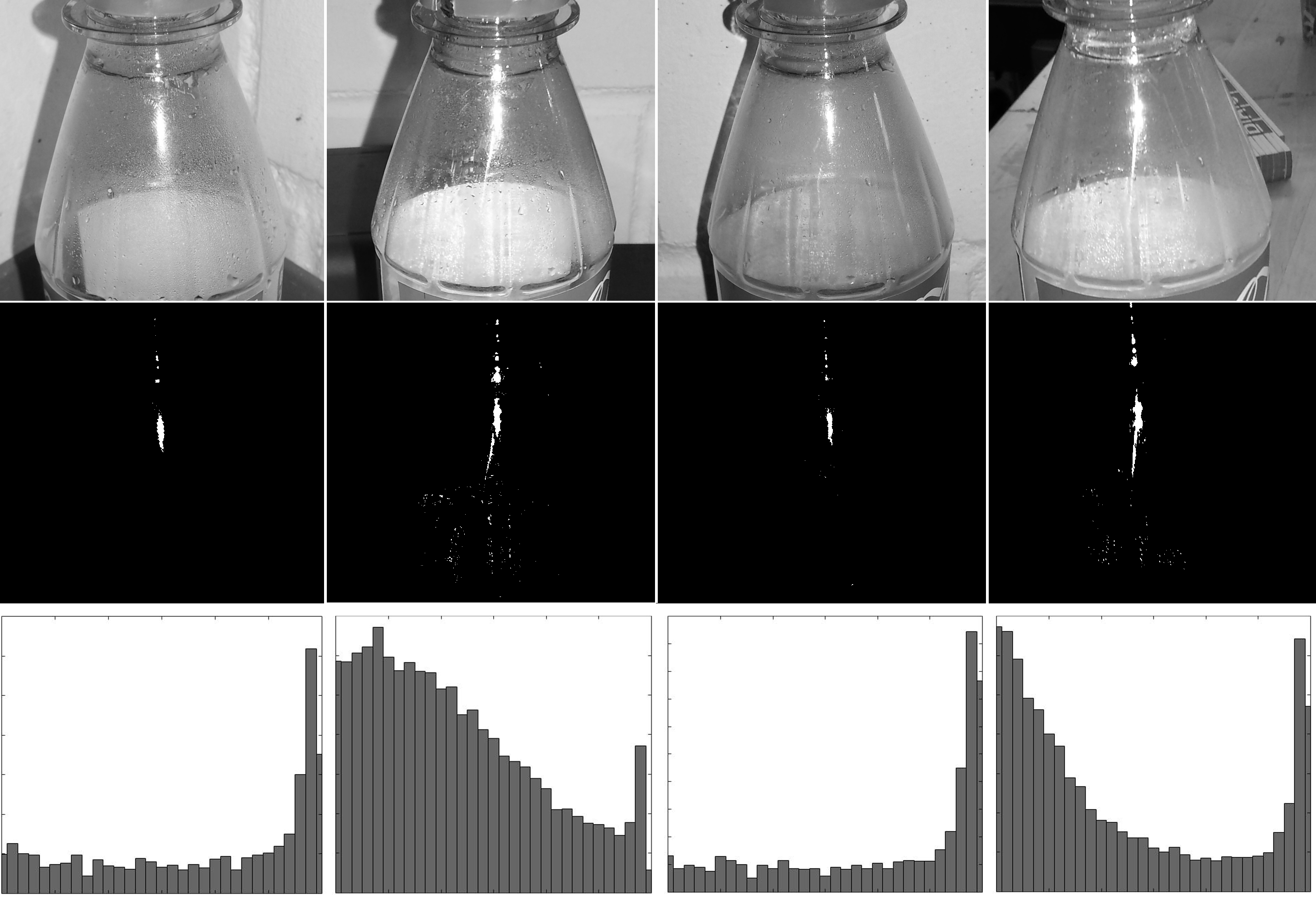}
	\caption[Lighting conditions]{Thresholded intensity images in various lighting environments. \textbf{Left to right}: dark corner of a sparsely lighted room to a bright sunlight situation. \textbf{Top}: desaturated images, \textbf{middle}: thresholded images, \textbf{bottom}: partial intensity distributions, with intensity as an 8 bit integer value. Showing the last 30 single value bins.}
	\label{lighting conditions}
\end{figure}

After reducing the RGB image to an intensity image by luminance based desaturation, pixels with intensity values beneath a certain threshold are discarded. The image now only contains highlight candidates as shown in the middle row of Figure \ref{intensity-specularities-y}. The highlights caused by the active light are mostly constant, except for situations having direct sunlight, as can be seen in Figure \ref{lighting conditions}. The intensity histograms below show a clear bump in the high range. By removing all pixels except for the last five intensity bins, we find a conservative compromise between capturing many (overshot) specular reflections whilst preserving robustness. The histograms do show most candidate specular pixels are covered by using these five bins. Obviously, the environment lighting does change the amount and position of highlights and thus produces false positives.

\begin{figure}[ht!]
\centering
	\includegraphics[width=85mm]{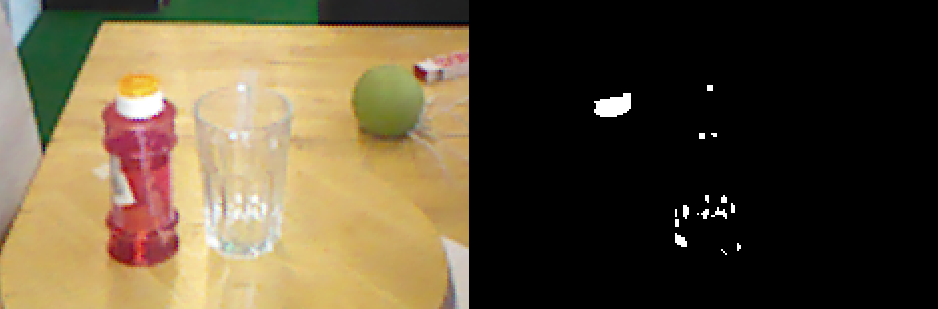}
	\caption[False positives specular highlight]{\textbf{Left}: original scene and \textbf{right}: specular highlight candidates. The white soap bottle cap produces a false positive.}
	\label{false positive}
\end{figure} 

Another source for false positive pixels are bright semi-Lambertian surfaces, as seen in Figure \ref{false positive}. As an alternative to the model based recognition done in \cite{Osadchy2003}, we only consider specular candidates that also do not show available depth data. We achieve more consistent results by using this crossmodal feature. Specular highlights are now represented as patches of high values in intensity space, allowing us to use the intensity gradient feature that was defined in the baseline system for $\mathcal{M}_1$.

\begin{figure}[ht!]
\centering
	\includegraphics[width=28mm]{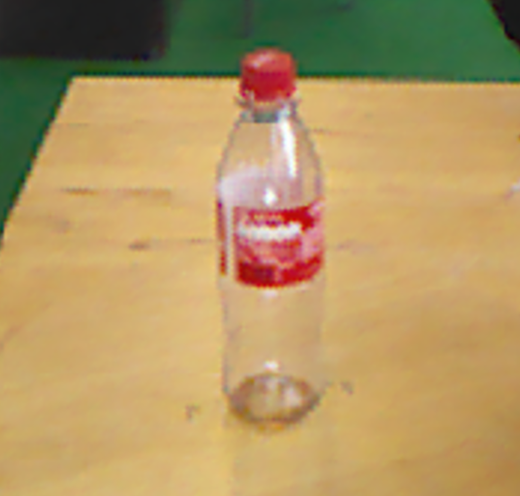}
	\includegraphics[width=28mm]{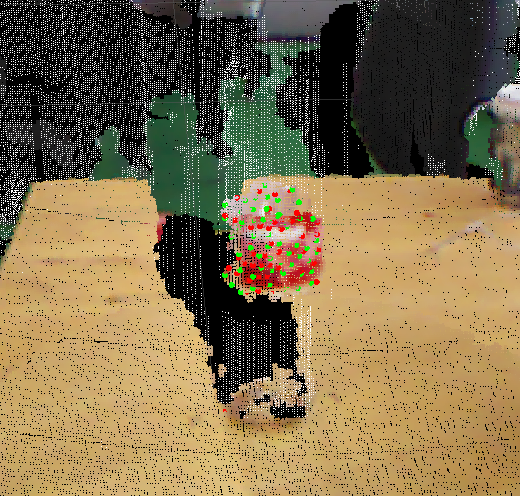}
	\includegraphics[width=28mm]{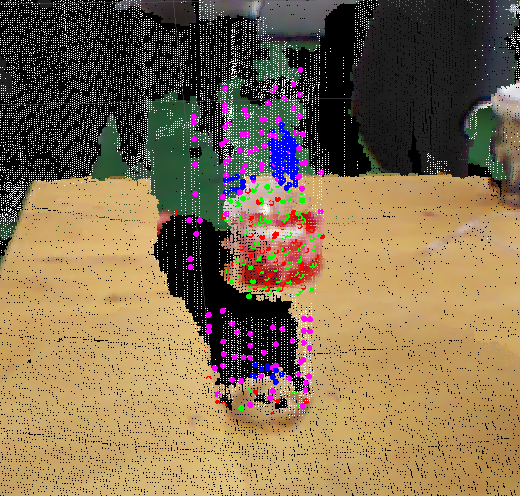}
	\caption[Feature comparison]{Comparison of two- and four-modal approach. \textbf{Left}: RGB image, \textbf{middle}: two-modal approach, \textbf{right}: extended approach with all four modalities. Intensity gradient features are represented in \textit{red}, normal vector features in \textit{green}, unavailable depth features in \textit{purple} and specular highlight features in \textit{blue}.}
	\label{comparison-features}
\end{figure}

As defined by Hinterstoisser et al. \cite{Hinterstoisser2012}, a template matrix with all 2D feature positions and their normalized cue values is constructed. Figure \ref{comparison-features} shows features' relative positions that are maintained in our extension, as opposed to the baseline. The coke bottle is a good example to show that the surface coverage of the four-modal approach is much bigger compared to the original two-modal approach. This helps to get a much better result in position estimation.
Another issue with the baseline approach is caused by it not being able to determine the relative position of the label on the object. It produces more false positives when other bottles are present, with labels positioned at different heights. On a final note, the new modalities tend to be complementary to the baseline features, promoting the idea of increased robustness by using complementary modalities from the baseline approach.

\begin{figure}[ht!]
\centering
	\includegraphics[width=85mm]{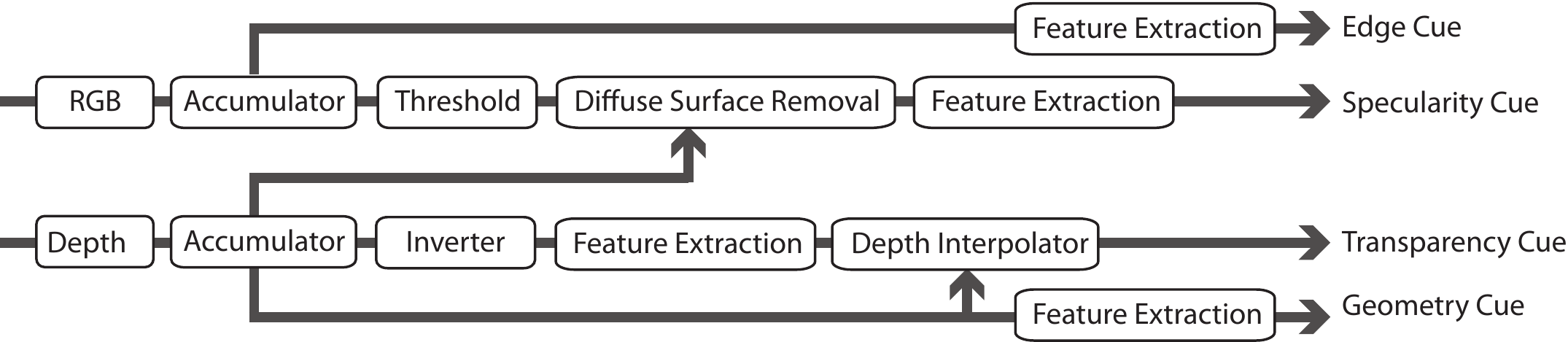}
	\caption{Modality pipelines}
	\label{architecture}
\end{figure}

Figure \ref{architecture} describes the pipelines that were used, whereby the \textit{Specularity Cue} and the \textit{Transparency Cue} belong to our extension.

Summarizing, we approach NIR pattern distortions by transparent objects and their specular reflections as binary patches, differing from the background, defined by its shape and position.

\subsection{Feature Stability}

\begin{figure}[ht!]
\centering
	\includegraphics[width=85mm]{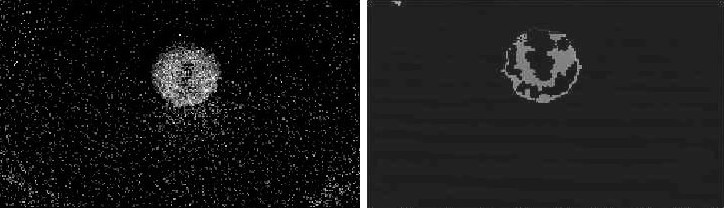}
	\caption[Unstable depth in NIR]{\textbf{Left}: maximum disparity fluctuation, represented as an intensity value. \textbf{Right}: the corresponding depth fluctuation.}
	\label{unstable-depth-ir}
\end{figure}

RGB-D (and stereo) cameras produce unstable depth maps, caused by refraction of NIR patterns by an object's surface. Unstable patches of pixels, as shown on the right of Figure \ref{unstable-depth-ir}, can invalidate a large subset of possible features, leading to a large amount of necessary templates per view.

On the other hand, our depth interpolation method, which is described in subsection \ref{sec:Localization} depends on the contour of the object being stable. Depth values tend to be very noisy on surface edges. We therefore use a per pixel running average to stabilize the depth data whilst minimizing the impact of preprocessing on the total processing time. By accumulating a number of frames and removing every pixel that reaches a \textit{nan} value at least once over all frames, we are able to reduce the amount of unstable pixels from 4\% to about 1\%. An accumulator was also used for RGB values. 

The specularity based modality is unstable in pose changes parallel to the camera plane. Figure \ref{intensity-specularities-y} however shows, that for small pose changes, they are stable enough to be used under the assumption that the dominant light direction is known.

\subsection{Localization}
\label{sec:Localization}

\begin{figure}[ht!]
\centering
	\includegraphics[width=120mm]{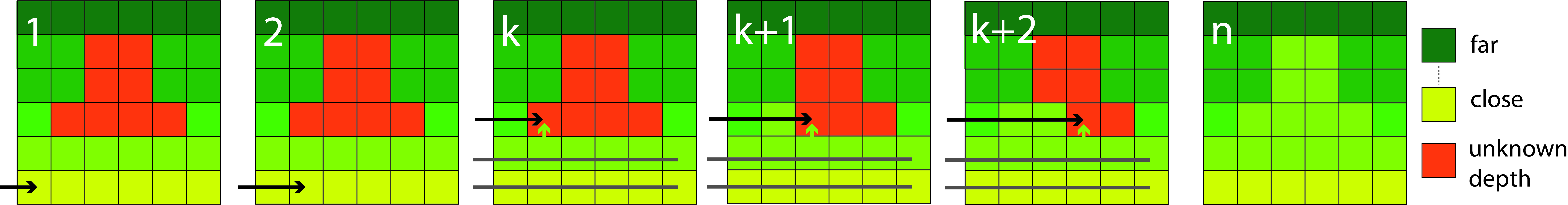}
	\caption[Walk down depth]{Scanline depth interpolation. We traverse the depth map bottom-up and from left to right. Once we find a \textit{nan} valued pixel, we traverse down the depth map and take the first valid depth value we find, pushing its value up into the unknown pixel. This algorithm depends on our pre-processing step, namely on removing most of the flickering noise that occurs especially on edges of objects.}
	\label{depth-interpolation-algorithm}
\end{figure}

Similar to the baseline approach, the recognition system returns the positions of the features as defined in the best-matching template. These positions can then be used to localize the object in 3D space, using both the depth map and the extrinsic parameters of the camera. As the object coverage is much higher for the extended approach when used on transparent objects, the localization quality is also higher. Depth value assignment to \textit{nan} valued pixels in the depth map is done using a scanline algorithm as depicted in Figure \ref{depth-interpolation-algorithm}, extruding depth values from the nearest horizontal surface beneath the object into the unknown depth values of the object. This approach is unique and stable enough to allow object recognition and enable the position estimation for transparent objects. In order to filter outlier pixels that occur especially around object edges (and jump between the object’s location and the
background), we remove these outliers using a statistical outlier filter, based on a Euclidean distance threshold. 

\section{Evaluation}\label{sec:Evaluation}

\begin{figure}[ht!]
\centering
	\includegraphics[width=75mm]{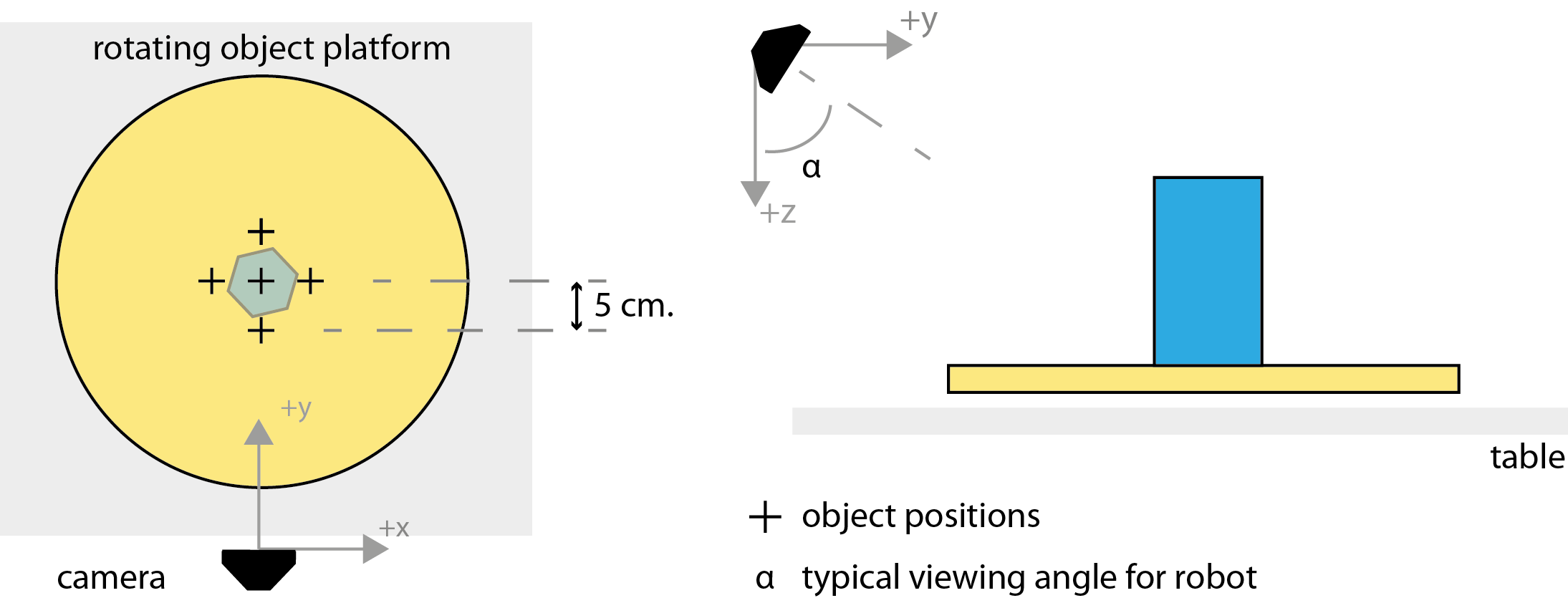}
	\includegraphics[width=35mm]{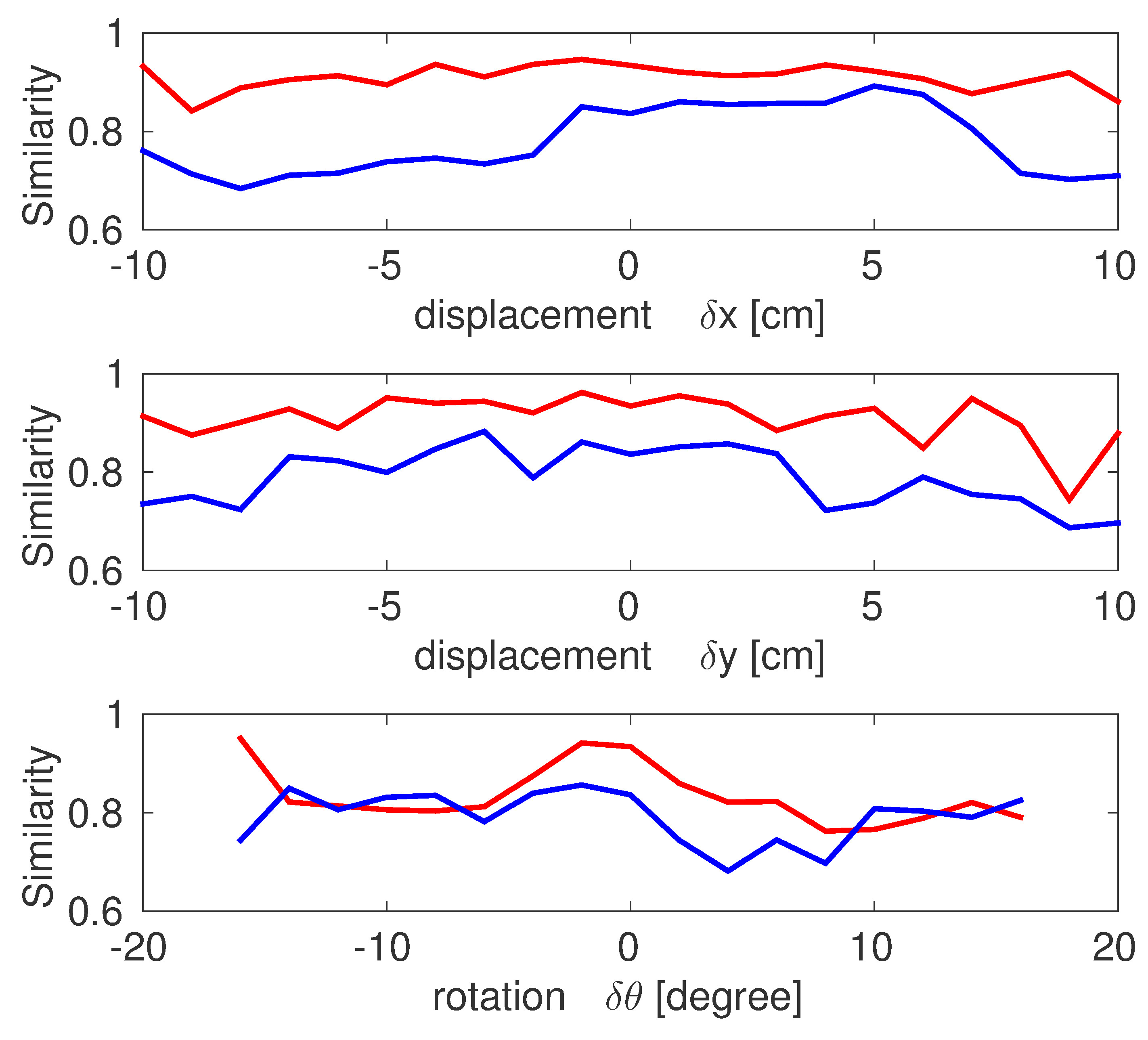}
	\caption{\textbf{Left}: main experiment setup. \textbf{Right}: similarity response while displacing/rotating the object. The experiment is based on one templated object pose, with red representing the dual modal approach and blue the 4-modal approach.}
	\label{experiment-setup}
\end{figure}

In order to evaluate the impact of the new modalities on the object recognition performance, we compare the original LineMOD feature set with three different setups, described in Table \ref{recognition_rates}. Tests are run on a Care-O-bot\textregistered\ 3 robot, equipped with a Kinect, using object poses found in common household scenarios, depicted in Figure \ref{experiment-setup}. We test nine objects from 3 different categories, three per category: \textit{diffuse}, \textit{transparent} and \textit{composite} materials. With $\mathcal{M}_1$ and $\mathcal{M}_2$ allowing recognition of objects from the first category, $\mathcal{M}_3$ and $\mathcal{M}_4$ are expected to work mostly on the second category and all modalities to work together on the latter. A robustness test against object pose changes and a full test on all objects is performed. In all experiments, true positives are defined as a template response that contains the correct object name and a 3D centroid location that is within 5 cm. from the ground truth object centroid location. All other responses are false positives.

In a first experiment, templates of a soda bottle are built for both the baseline as well as the fully extended modality setups simultaneously, after which the template similarity values are extracted from the scene at varying object poses. Figure \ref{experiment-setup} (right) shows similarity responses whilst changing $x$, $y$ and $theta$ object coordinates. A single template shows similar behaviour for the extended approach, although a dip for $x$ (parallel to the camera) movement can be seen, which can be explained by the bottle's label center not pointing towards the camera. Similarity values in general are lower for the 4-modal approach, which produces about twice as many features per object, whereby $\mathcal{M}_3$'s features are less stable, because of the instability of the interpolated depth values, and thus produces a lower similarity level.

In the second experiment, templates are created for 1000 different object poses (5 positions, depicted in Figure \ref{experiment-setup}, each from 200 rotated views). A template is only created when the database does not already contain a matching one, preventing duplicate templates. Table \ref{experiment3-table-nrfeatures-templates} shows the actual created number of templates per object, showing that the number of needed templates is greatly reduced when using a larger number of modalities.

\begin{table}[ht!]
\centering
	\caption{Total number of templates per object for all four modality combinations}
	\begin{tabular}{|l|l|l|l|l|l|}
		\hline 
	 	Set&$\mathcal{M}_1$, $\mathcal{M}_2$  	&	$\mathcal{M}_1$, $\mathcal{M}_2$, $\mathcal{M}_3$  	&	$\mathcal{M}_1$, $\mathcal{M}_2$, $\mathcal{M}_4$ 	& all \\
	 	\hline 
		Large noodles 	&309  		&60  		&256  	&79 \\
		\hline
		Small noodles 	&358  		&87  		&286  	&108 \\
		\hline 
		Candle 		&345  		&67  		&273  	&85 \\
		\hline 
		Coke bottle 	&382  		&225  	&399  &229 \\
		\hline 
		Sprite bottle 	&92  		&19  		&53  	&23 \\
		\hline 
		Soap bubbles 	&98  		&26  		&72  	&33 \\
		\hline 
		Water glass 	&262  		&33  		&169  &48 \\
		\hline 
		Beer glass 	&300  		&48  		&249  &68 \\
		\hline 
		Wine glass 	&83  		&11  		&44  	&13 \\
		\hline 
	\end{tabular}
	\label{experiment3-table-nrfeatures-templates}
\end{table}

\begin{figure}[ht!]
\centering
	\includegraphics[width=42mm]{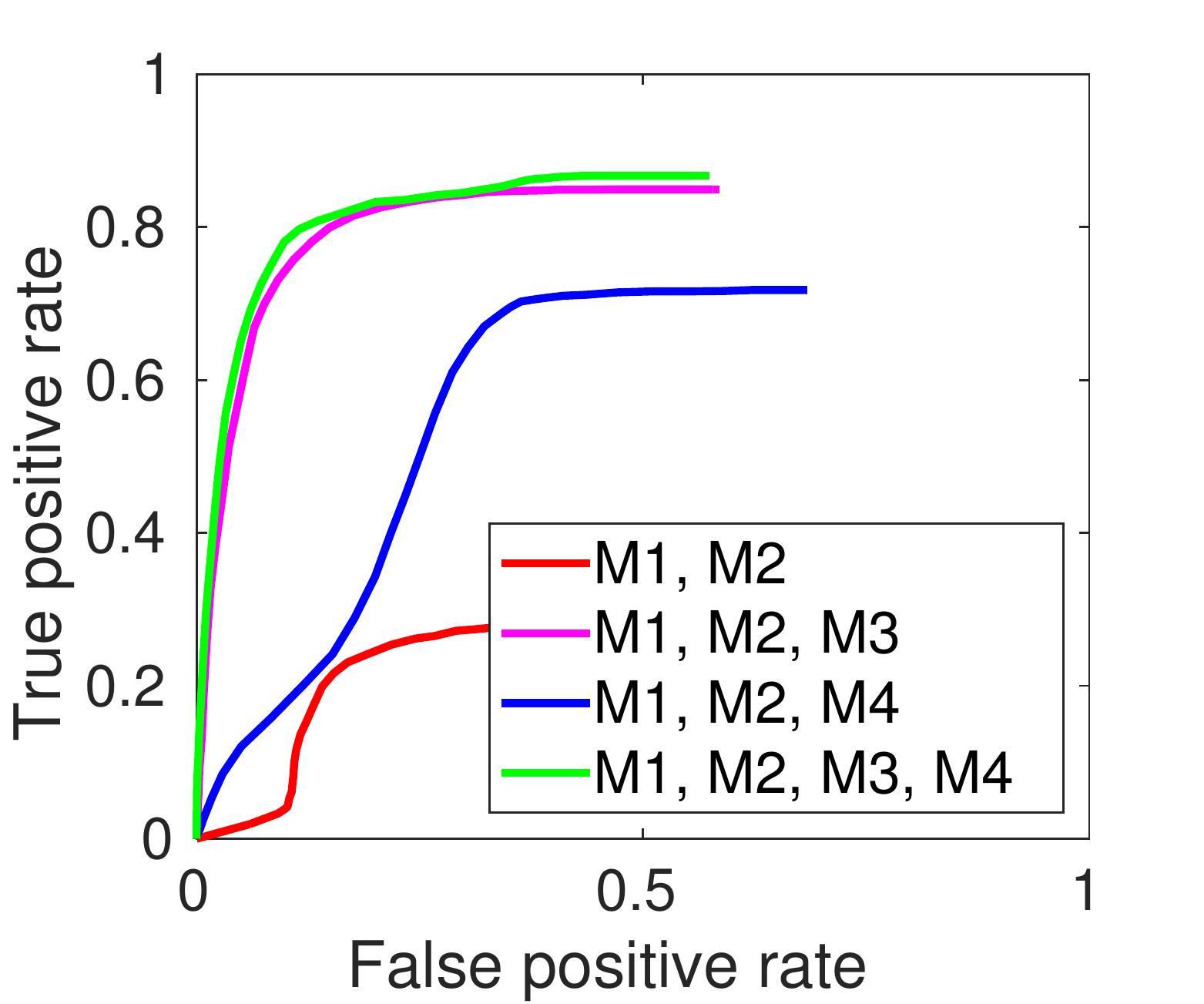}
	\includegraphics[width=42mm]{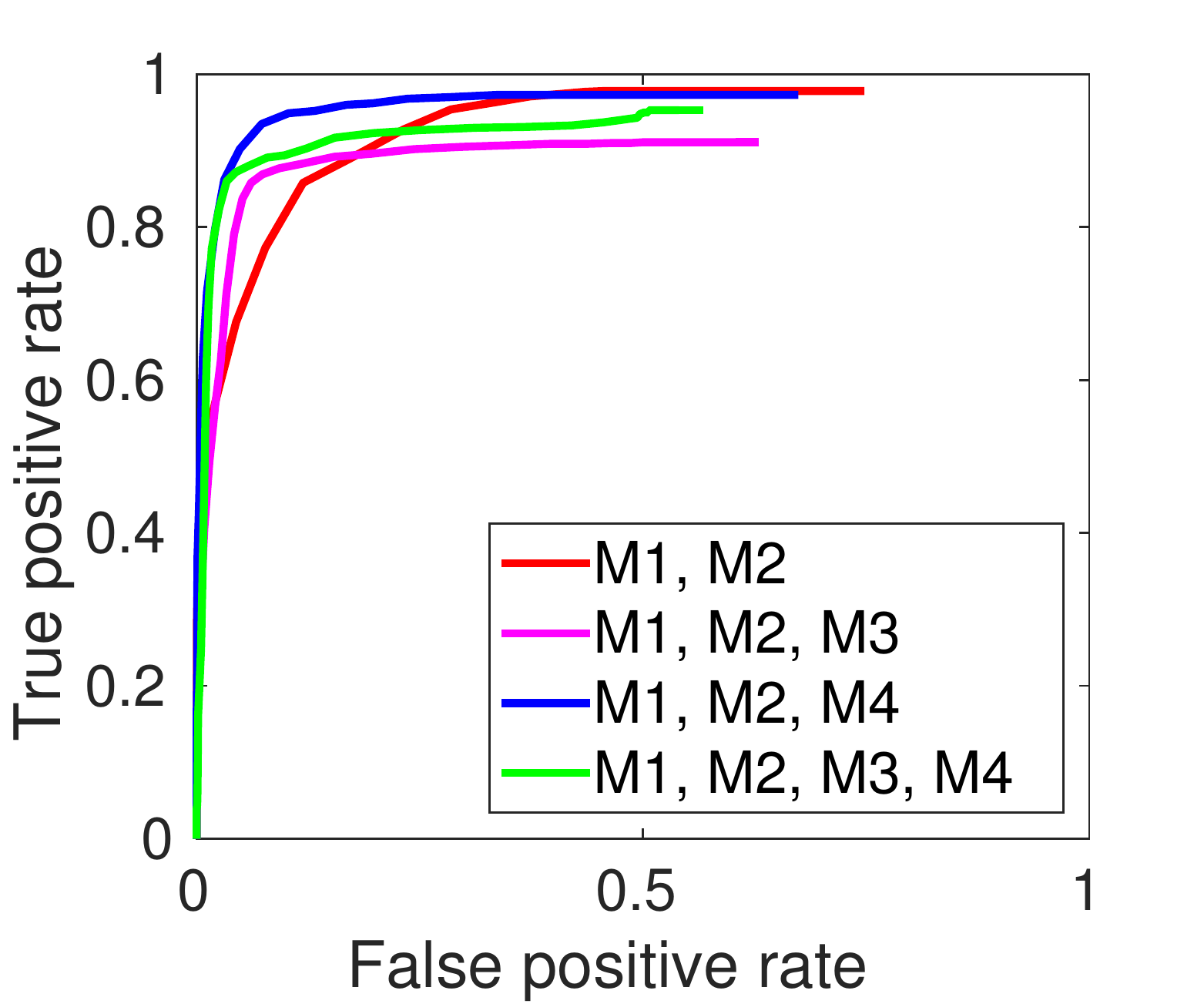}
		\caption[Main experiment]{\textbf{Left}: performance on the entire object set. \textbf{Right}: performance on the reduced set, not containing any transparent objects.}
	\label{experiment-full}
\end{figure}

The trained database is used in 250 recognition trials per object, which are performed with randomly generated object poses (on the rotating platform). Figure \ref{experiment-full} shows the ROC curves on the entire object set for all four systems. The similarity threshold, which is used to decide whether a returned similarity response is counted as a positive result, is varied between 0 and 100 percent. The extension shows saturation at around 92\% because the features from $\mathcal{M}_3$ are not completely stable, producing some mismatches between the templates and the scene. 

Adding any of the newly proposed modalities greatly improves the results, whereby the combination of all four produces the best results, although the system is never able to recognize all objects. Nevertheless, the modality combination seems a step into the right direction. The unknown depth modality introduces the largest improvement, which can be explained by the much higher object surface coverage that is reached with this modality. Recognition rates are shown in Table \ref{recognition_rates}, showing a significant improvement. The right of Figure \ref{experiment-full} shows the results of a reduced experiment with all transparent objects removed from the object set, showing that even for \textit{diffuse} objects, our extension improves sensitivity and thus robustness. Table \ref{recognition_rates} shows the resulting recognition rates, when picking the optimal similarity response decision boundary, which we found to be 75\%, which is a bit lower than the boundary found by Hinterstoisser et al. The recognition rates on the diffuse-only object set are similar for all modality combinations, except using $\mathcal{M}_3$, which can be explained by the added edge noise, for the NIR light is deflected from those edges.

\begin{table}[ht!]
\centering
	\caption[Recognition rates]{Recognition rates}
	\label{recognition_rates}
	\begin{tabular}{|l|l|l|l|l|l|}
		\hline 
	 	Set&$\mathcal{M}_1$, $\mathcal{M}_2$  	&	$\mathcal{M}_1$, $\mathcal{M}_2$, $\mathcal{M}_3$  	&	$\mathcal{M}_1$, $\mathcal{M}_2$, $\mathcal{M}_4$ 	& all \\
	 	\hline 
		 	all	 &28\%  		&79\%  		&71\%  		&\textbf{81\%} \\
		\hline
		 	diffuse	 &\textbf{98\%}  		&89\%  		&97\%  		&92\% \\
		\hline 
	\end{tabular}
\end{table}

Table \ref{experiment3-table-timings} shows the average total time needed for a full database comparison, showing an approximate doubling of recognition time, caused by the processing of the added modalities.

\begin{table}[ht!]
\centering
	\caption[Experiment 3 timings]{Average time needed for template comparison using all templates.}
	\begin{tabular}{|l|l|l|l|l|l|}
		\hline 
	 	Set&$\mathcal{M}_1$, $\mathcal{M}_2$  	&	$\mathcal{M}_1$, $\mathcal{M}_2$, $\mathcal{M}_3$  	&	$\mathcal{M}_1$, $\mathcal{M}_2$, $\mathcal{M}_4$ 	& all \\
	 	\hline 
		Time [seconds] 	&0.031  		&0.058  		&0.064  		&0.065 \\
		\hline
	\end{tabular}
	\label{experiment3-table-timings}
\end{table}

\section{Conclusion}\label{sec:Conclusion}

We have shown that a combination of multiple modalities, designed for various physical aspects of materials, leads to an increase in robustness for object recognition. By reinterpreting color and depth data, we are able to distinguish objects from diffuse, transparent and composite categories. It improves recognition of these objects without significantly reducing the recognition rate and time of diffuse objects. The recognition rate is improved most significantly by use of an \textit{unavailable depth} modality.

In future work, we will utilize NIR active lighting to find the specular reflection response in order to minimize false positives that are produced by unknown external light sources. A new modality will be added based on the fact that geometry, especially edges, causes noise in the depth map. This can serve as a cue for object borders.

Using a larger number of sensor modalities and making use of the weaknesses of low cost sensors allows to recognize objects from a large amount of material categories. For robotics, solving object recognition in an affordable, generalizable and robust way is of utmost importance, as it will increase the acceptance rate of users in the general public.


\bibliographystyle{splncs03}
\bibliography{maindocument.bib}
\end{document}